\documentclass[11pt,a4paper]{article}
\usepackage[hyperref]{acl2018}
\usepackage{times}
\usepackage{latexsym}

\usepackage{url}
\usepackage{xcolor}
\usepackage{soul}
\usepackage[utf8]{inputenc}

\usepackage{CJKutf8}
\usepackage{amsmath}
\usepackage{graphicx}
\usepackage{multirow}
\usepackage[linesnumbered,ruled,vlined]{algorithm2e}
\usepackage{url}
\usepackage{makecell}

\usepackage{amsfonts}
\usepackage{colortbl}
\usepackage{color,framed}
\definecolor{mygray}{gray}{.9}
\definecolor{mypink}{rgb}{.99,.91,.95}
\definecolor{mycyan}{cmyk}{.3,0,0,0}
\aclfinalcopy 

\title{Deep Reinforcement Learning for Chinese Zero Pronoun Resolution}

\author{Qingyu Yin$^\sharp$, Yu Zhang$^\sharp$, Weinan Zhang$^\sharp$, Ting Liu$^\sharp$\thanks{\ \ Corresponding author.}, William Yang Wang$^\flat$ \\
  $^\sharp$Harbin Institute of Technology, China \\
  $^\flat$University of California, Santa Barbara, USA\\
  {\tt \{qyyin, yzhang, wnzhang, tliu\}@ir.hit.edu.cn} \\
  {\tt william@cs.ucsb.edu} \\
  }

\date{}
\begin{document}
\begin{CJK*}{UTF8}{gbsn}
\maketitle
\begin{abstract}
Deep neural network models for Chinese zero pronoun resolution learn semantic information for zero pronoun and candidate antecedents, but tend to be short-sighted---they often make local decisions. They typically predict coreference chains between the zero pronoun and one single candidate antecedent one link at a time, while overlooking their long-term influence on future decisions. Ideally, modeling useful information of preceding potential antecedents is critical when later predicting zero pronoun-candidate antecedent pairs. In this study, we show how to integrate local and global decision-making by exploiting deep reinforcement learning models. With the help of the reinforcement learning agent, our model learns the policy of selecting antecedents in a sequential manner, where useful information provided by earlier predicted antecedents could be utilized for making later coreference decisions. Experimental results on OntoNotes 5.0 dataset show that our technique surpasses the state-of-the-art models.
\end{abstract}

\section{Introduction} 
\label{sec:introduction}
Zero pronoun, as a special linguistic phenomenon in pro-dropped languages, is pervasive in Chinese documents~\cite{zhao2007}. A zero pronoun is a gap in the sentence, which refers to the component that is omitted because of the coherence of language. Following shows an example of zero pronoun in Chinese document, where zero pronouns are represented as ``$\phi$''.
        \begin{quote}\small
            [当事人\ 李亚鼎] 除了\ 表示\ {\bf $\phi_1$ }欣然\ 接受\ 但\  {\bf $\phi_2$ } 也\ 希望\ 国家\ 要\ 有\ 人\ 负责 。

            ([Litigant Li Yading] not only shows {\bf $\phi_1$} willing of acception, but also {\bf $\phi_2$} hopes that there should be someone in charge of it.) 

        \end{quote}
A zero pronoun can be an anaphoric zero pronoun if it coreferes to one or more mentions in the associated text, or unanaphoric, if there are no such mentions. In this example, the second zero pronoun ``{\bf $\phi_2$}'' is anaphoric and corefers to the mention ``当事人 李亚鼎/Litigant Li Yading'' while the zero pronoun ``{\bf $\phi_1$}'' is unanaphoric. These mentions that contain the important information for interpreting the zero pronoun are called the antecedents.

In recent years, deep learning models for Chinese zero pronoun resolution have been widely investigated~\cite{chen2016,yin2017chinese,yin2017chinesezp}. These solutions concentrate on anaphoric zero pronoun resolution, applying numerous neural network models to zero pronoun-candidate antecedent prediction. Neural network models have demonstrated their capabilities to learn vector-space semantics of zero pronouns and their antecedents~\cite{yin2017chinese,yin2017chinesezp}, and substantially surpass classic models~\cite{zhao2007,chen2013,chen2015}, obtaining state-of-the-art results on the benchmark dataset.

However, these models are heavily making local coreference decisions. They simply consider the coreference chain between the zero pronoun and one single candidate antecedent one link at a time while overlooking their impacts on future decisions. Intuitively, antecedents provide key linguistic cues for explaining the zero pronoun, it is therefore reasonable to leverage useful information provided by previously predicted antecedents as cues for predicting the later zero pronoun-candidate antecedent pairs. For instance, given a sentence ``I have confidence that $\phi$ can do it.'' with its candidate mentions ``he'', ``the boy'' and ``I'', it is challenging to infer whether mention ``I'' is possible to be the antecedent if it is considered separately. In that case, the resolver may incorrectly predict ``I'' to be the antecedent since ``I'' is the nearest mention. Nevertheless, if we know that ``he'' and ``the boy'' have already been predicted to be the antecedents, it is uncomplicated to infer the $\phi$-``I'' pair as ``non-coreference'' because ``I'' corefers to the disparate entity that is refered by ``he''. Hence, a desirable resolver should be able to 1) take advantage of cues of previously predicted antecedents, which could be incorporated to help classify later candidate antecedents and 2) model the long-term influence of the single coreference decision in a sequential manner.

To achieve these goals, we propose a deep reinforcement learning model for anaphoric zero pronoun resolution. On top of the neural network models~\cite{yin2017chinese,yin2017chinesezp}, two main innovations are introduced that are capable of efficaciously leveraging effective information provided by potential antecedents, and making long-term decisions from a global perspective. First, when dealing with a specific zero pronoun-candidate antecedent pair, our system encodes all its preceding candidate antecedents that are predicted to be the antecedents in the vector space. Consequently, this representative vector is regarded as the antecedent information, which can be utilized to measure the coreference probability of the zero pronoun-candidate antecedent pair. In addition, the policy-based deep reinforcement learning algorithm is applied to learn the policy of making coreference decisions for zero pronoun-candidate antecedent pairs. The innovative idea behind our reinforcement learning model is to model the antecedent determination as a sequential decision process, where our model learns to link the zero pronoun to its potential antecedents incrementally. By encoding the antecedents predicted in previous states, our model is capable of exploring the long-term influence of independent decisions, producing more accurate results than models that simply consider the limited information in one single state.

Our strategy is favorable in the following aspects. First, the proposed model learns to make decisions by linguistic cues of previously predicted antecedents. Instead of simply making local decisions, our technique allows the model to learn which action (predict to be an antecedent) available from the current state can eventually lead to a high-scoring overall performance. Second, instead of requiring supervised signals at each time step, deep reinforcement learning model optimizes its policy based on an overall reward signal. In other words, it learns to directly optimize the overall evaluation metrics, which is more effective than models that learn with loss functions that heuristically define the goodness of a particular single decision. Our experiments are conducted on the OntoNotes dataset. Comparing to baseline systems, our model obtains significant improvements, achieving the state-of-the-art performance for zero pronoun resolution. The major contributions of this paper are three-fold.

\begin{itemize}
\item We are the first to consider reinforcement learning models for zero pronoun resolution in Chinese documents;

\item The proposed deep reinforcement learning model leverages linguistic cues provided by the antecedents predicted in earlier states when classifying later candidate antecedents;

\item We evaluate our reinforcement learning model on a benchmark dataset, where a considerable improvement is gained over the state-of-the-art systems.
\end{itemize}

The rest of this paper is organized as follows. The next section describes our deep reinforcement learning model for anaphoric zero pronoun resolution. Section \ref{sec:experiments} presents our experiments, including the dataset description, evaluation metrics, experiment results, and analysis. We outline related work in Section \ref{sec:related_work}. The Section \ref{sec:conclude} is about the conclusion and future work.

        \begin{figure*}
            \centering
            \includegraphics[width=0.9\textwidth]{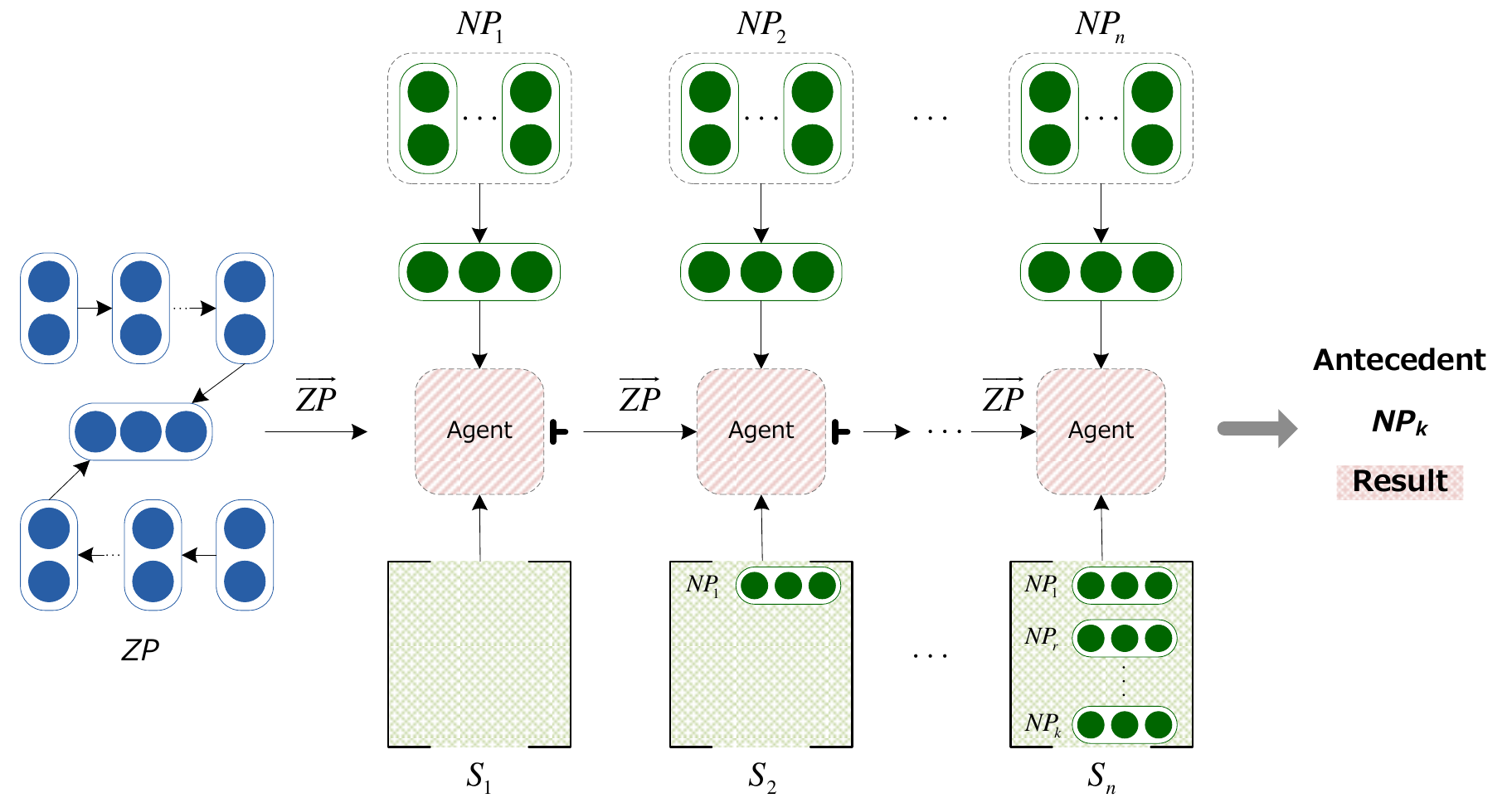}
            \caption{Illustration of our reinforcement learning framework. Given a zero pronoun with $n$ candidate antecedents (presented as ``NP''), for each time, the agent scores pairs of zero pronoun-candidate antecedent for their likelihood of coreference by 1) zero pronoun; 2) candidate antecedent and 3) antecedent information. Antecedent information at time $t$ is generated by all the antecedents predicted in previous states.}
            \label{all}
        \end{figure*}

\section{modelology} 
\label{sec:model} 
In this section, we introduce the technical details of the proposed reinforcement learning framework. The specific task of anaphoric zero pronoun resolution is to select antecedents from candidate antecedents for the zero pronoun. Here we formulate it as a sequential decision process in a reinforcement learning setting. We first describe the environment of the Markov decision making process and our reinforcement learning agent. Then, we introduce the modules. The last subsection is about the supervised pre-training technique of our model.

\subsection{Reinforcement Learning for Zero Pronoun Resolution}

Given an anaphoric zero pronoun $zp$, a set of candidate antecedents are required to be selected from its associated text. In particular, we adopt the heuristic model utilized in recent Chinese anaphoric zero pronoun resolution work~\cite{chen2016,yin2017chinese,yin2017chinesezp} for this purpose. For those noun phrases that are two sentences away at most from the zero pronoun, we select those who are maximal noun phrases or modifier ones to compose the candidate set. These noun phrases ($\{np_1, np_2,..., np_n\}$) and the zero pronoun ($zp$) are then encoded into representation vectors: $\{v_{np_1}, v_{np_2},..., v_{np_n}\}$ and $v_{zp}$.

Previous neural network models~\cite{chen2016,yin2017chinese,yin2017chinesezp} generally consider some pairwise models to select antecedents. In these work, candidate antecedents and the zero pronoun are first merged into pairs $\{(zp,np_1), (zp,np_2), ..., (zp,np_n)\}$, and then different neural networks are applied to deal with each pair independently. We argue that these models only make local decisions while overlooking their impacts on future decisions. In contrast, we formulate the antecedent determination process in as Markov decision process problem. An innovative reinforcement learning algorithm is designed that learns to classify candidate antecedents incrementally. When predicting one single zero pronoun-candidate antecedent pair, our model leverages antecedent information generated by previously predicted antecedents, making coreference decisions based on global signals.

The architecture of our reinforcement learning framework is shown in Figure~\ref{all}. For each time step, our reinforcement learning agent predicts the zero pronoun-candidate antecedent pair by using 1) the zero pronoun; 2) information of current candidate antecedent and 3) antecedent information generated by antecedents predicted in previous states. In particular, our reinforcement learning agent is designed as a policy network $\pi_\theta(s,a) = p(a|s;\theta)$, where $s$ represents the \emph{state}; $a$ indicates the \emph{action} and $\theta$ represents the parameters of the model. The parameters $\theta$ are trained using stochastic gradient descent. Compared with Deep Q-Network \cite{mnih2013playing} that commonly learns a greedy policy, policy network is able to learn a stochastic policy that prevents the agent from getting stuck at an intermediate state \cite{xiong2017deeppath}. Additionally, the learned policy is more explainable, comparing to learned value functions in Deep Q-Network. We here introduce the definitions of components of our reinforcement learning model, namely, \emph{state}, \emph{action} and \emph{reward}.

\subsubsection{State}

Given a zero pronoun $zp$ with its representation $v_{zp}$ and all of its candidate antecedents representations $\{v_{np_1}, v_{np_2} ,..., v_{np_n}\}$, our model generate coreference decisions for zero pronoun-candidate antecedent pairs in sequence. More specifically, for each time, the \emph{state} is generated by using both the vectors of the current zero pronoun-candidate antecedent pair and candidates that have been predicted to be the antecedents in the previous states. For time $t$, the state vector $s_t$ is generated as follows:
\begin{equation}
        s_t = (v_{zp},v_{np_t},v_{ante}(t), v_{feature_t})
\end{equation}
where $v_{zp}$ and $v_{np_t}$ are the vectors of $zp$ and $np_t$ at time $t$. As shown in \newcite{chen2016}, human-designed handcrafted features are essential for the resolver since they reveal the syntactical, positional and other relations between a zero pronoun and its counterpart antecedents. Hence, to evaluate the coreference possibility of each candidate antecedent in a comprehensive manner, we integrate a group of features that are utilized in previous work \cite{zhao2007,chen2013,chen2016} into our model.  For these multi-value features, we decompose them into a corresponding set of binary-value ones. $v_{feature_t}$ represents the feature vector. $v_{ante}(t)$ represents the antecedent information generated by candidates that have been predicted to be antecedents in previous states. After that, these vectors are concatenated to be the representation of \emph{state} and fed into the deep reinforcement learning agent to generate the \emph{action}.

\subsubsection{Action}
The action for each state is defined to be: \emph{corefer} that indicates the zero pronoun and candidate antecedent are coreference; or otherwise, \emph{non-corefer}. If an action \emph{corefer} is made, we retain the vector of the counterpart antecedent together with those of the antecedents predicted in previous states to generate the vector $v_{ante}$, which is utilized to produce the antecedent information in the next state.

\subsubsection{Reward}
Normally, once the agent executes a series of actions, it observes a reward $R(a_{1:T} )$ that could be any function. To encourage the agent to find accurate antecedents, we regard the F-score for the selected antecedents as the \emph{reward} for each action in a path.

\subsection{Reinforcement Learning Agent}
Basically, our reinforcement learning agent is comprised of three parts, namely, the zero pronoun encoder that learns to encode a zero pronoun into vectors by using its context words; the candidate mention encoder that represents the candidate antecedents by content words; and the agent that maps the state vector $s$ to a probability distribution over all possible actions.

In this work, the ZP-centered neural network model proposed by \newcite{yin2017chinese} is employed to be the zero pronoun encoder. The encoder learns to encode the zero pronoun by its associated text into its vector-space semantics. In particular, two standard recurrent neural networks are employed to encode the preceding text and the following text of a zero pronoun, separately. Such a model learns to encode the associated text around the zero pronoun, exploiting sentence-level information for the zero pronoun. For the candidate mentions encoder, we adopt the recurrent neural network-based model that encodes these phrases by using its content words. More specifically, we utilize a standard recurrent neural network to model the content of a phrase from left to right. This model learns to produce the vector of a phrase by considering its content, providing our model an ability to reveal its vector-space semantics. In this way, we generate the vector for $zp$, the $v_{zp}$, and representation vectors of all its candidate antecedents, which are denoted as $\{v_{np_1}, v_{np_2} ,..., v_{np_n}\}$.

Moreover, we employ pooling operations to encode antecedent information by using the antecedents that are predicted in previous states. In particular, we generate two vectors by applying the max-pooling and average-pooling, respectively. These two vectors are then concatenated together. Let the representative vector of the $t$th candidate antecedent to be $v_{np_t} \in\mathbb{R}^{d}$, and the predicted antecedents at time $t$ be writen as $S(t) = [v_{np_i},v_{np_j},...,v_{np_r}]$, the vector at time $t$,  $v_{ante}(t)_k$ is generated by:
$$v_{ante}(t)_k=
\begin{cases}
max\{S(t)_{k,\cdot}\}& \text{for 0 $\le$ k $<$ d}\\
ave\{S(t)_{k-d,\cdot}\}& \text{for d $\le$ k $<$ 2d}
\end{cases}$$

\begin{figure}[t]
            \centering
            \includegraphics[width=0.45\textwidth]{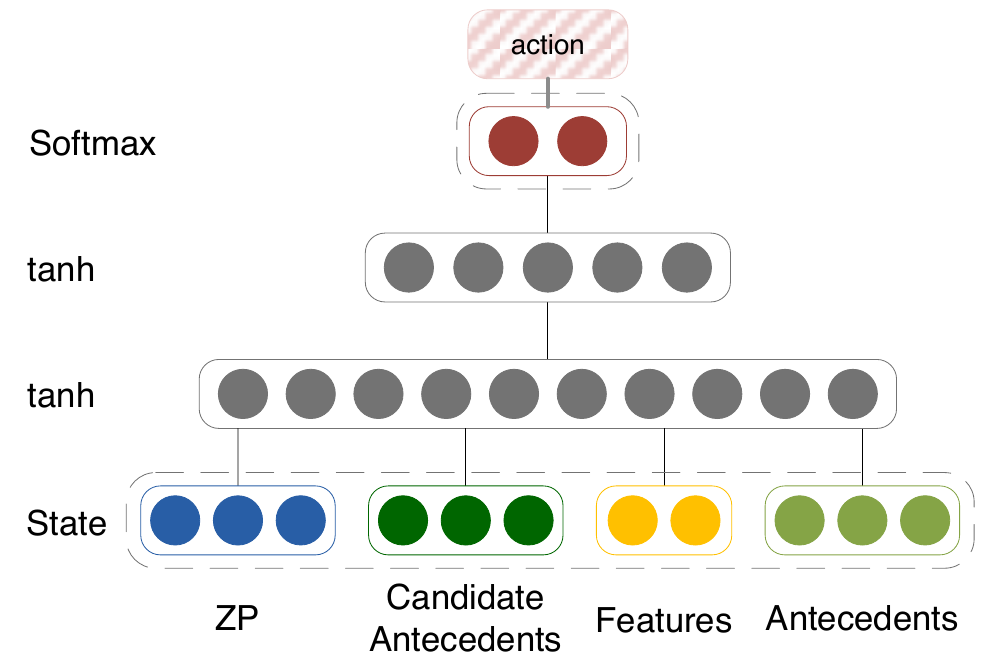}
            \caption{Illustration of the feedforward neural network model employed as the agent.
            Its input vector includes these parts: (1) Zero pronoun; (2) Candidate Antecedents; (3) Pair Features and (4) Antecedents.
            By going through all the full-connected hidden layers and one $softmax$ layer, the agent maps the state vector into the probability distribution over actions that indicates the coreference likelihood of the input zero pronoun-candidate antecedent pair.}
            \label{agent}
        \end{figure}

The concatenation of these vectors is regarded as input and is fed into our reinforcement learning agent. More specifically, a feed-forward neural network is utilized to constitute the agent that maps the state vector to a probability distribution over all possible actions. Figure~\ref{agent} shows the architecture of the agent.
Two hidden layers are employed in our model, each of which utilizes the $tanh$ as the activation function. For each layer, we generate the output by:
\begin{equation}
        h_i(s_t) = tanh(W_ih_{i-1}(s_t)+b_i)
\end{equation}
where $W_i$ and $b_i$ are the parameters of the $i$th hidden layer; $s_i$ represents the state vector. After going through all the layers, we can get the representative vector for the zero pronoun-candidate antecedent pair $(zp,np_t)$. We then feed it into a scoring-layer to get their coreference score. The scoring-layer is a fully-connected layer of dimension $2$:
\begin{equation}
        score(zp,np_t) = W_sh_{2}(s_t)+b_s
\end{equation}
where $h_{2}$ represents the output of the second hidden layer; $W_s\in\mathbb{R}^{2\times r}$ is the parameter of the layer and $r$ is the dimension of $h_{2}$. Consequently, we generate the probability distribution over actions using the output generated by the scoring-layer of the neural network, where a $softmax$ layer is employed to gain the probability of each action:
\begin{equation}
        p_\theta(a) \propto e^{score(zp,np_t)}
\end{equation}
In this work, the policy-based reinforcement learning model is employed to train the parameter of the agent. More specifically, we explore using the REINFORCE policy gradient algorithm~\cite{williams1992simple}, which learns to maximize the expected reward:
\begin{equation}
\begin{split}
        J(\theta) = \mathbb{E}_{a_{1:T}\sim p(a|zp,np_t;\theta)}R(a_{1:T})\ \ \ \ \ \ \ \ \ \\
        = \sum_t\sum_a p(a|zp,np_t;\theta)R(a_t)
        \end{split}
\end{equation}
where $p(a|zp,np_t;\theta)$ indicates the probability of selecting action $a$.

Intuitively, the estimation of the gradient might have very high variance. One commonly used remedy to reduce the variance is to subtract a \emph{baseline} value $b$ from the reward. Hence, we utilize the gradient estimate as follows:
\begin{equation}
        \nabla_{\theta}J(\theta) =  \nabla_{\theta} \sum_t \log p(a|zp,np_t;\theta)(R(a_{t})-b_t) 
\end{equation}
Following~\newcite{clark2016deep}, we intorduce the {baseline} $b$ and get the value of $b_t$ at time $t$ by $\mathbb{E}_{a_{t'}\sim p}R(a_1,...,a_{t'},...,a_T)$.

\subsection{Pretraining}
Pretraining is crucial in reinforcement learning techniques \cite{clark2016deep,xiong2017deeppath}. In this work, we pretrain the model by using the loss function from \newcite{yin2017chinese}:
     \begin{equation}
            loss = -\sum_{i = 1}^{N}\sum_{np \in \mathcal{A}(zp_i)}\delta(zp_i,np)log(P(np | zp_i))
    \end{equation}
where $P(np | zp_i)$ is the coreference score generated by the agent (the probability of choosing \emph{corefer} action); $\mathcal{A}(zp_i)$ represents the candidate antecedents of $zp_i$; $\delta(zp,np)$ is $1$ or $0$, representing $zp$ and $np$ are coreference or not.

\section{Experiments}
\label{sec:experiments}

\subsection{Dataset and Settings}
\subsubsection{Dataset}
Same to recent work on Chinese zero pronoun \cite{chen2016,yin2017chinese,yin2017chinesezp}, the proposed model is evaluated on the Chinese portion of the OntoNotes 5.0 dataset\footnote{\url{http://catalog.ldc.upenn.edu/LDC2013T19}} that was used in the Conll-2012 Shared Task. Documents in this dataset are from six different sources, namely, Broadcast News ($BN$), Newswires ($NW$), Broadcast Conversations ($BC$), Telephone Conversations ($TC$), Web Blogs ($WB$) and Magazines ($MZ$). Since zero pronoun coreference annotations exist in only the training and development set~\cite{chen2016}, we utilize the training dataset for training purposes and test our model on the development set. The statistics of our dataset are reported in Table~\ref{files}. To make equal comparison, we adopt the strategy as utilized in the existing work~\cite{chen2016,yin2017chinese}, where 20\% of the training dataset are randomly selected and reserved as a development dataset for tuning the model.
       \begin{table}[h]
        \begin{center}
        \begin{tabular}{lcccc}
        \Xhline{1pt}
        & $\#$Documents & $\#$Sentences & $\#$AZPs \\
        \hline
        Training & 1,391 & 36,487 & 12,111\\
        Test & 172 & 6,083 &  1,713\\
        \Xhline{1pt}
        \end{tabular}
        \end{center}
        \caption{Statistics on the training and test dataset.}
        \label{files}
        \end{table}
\subsubsection{Evaluation Measures}
Following previous work on zero pronoun resolution~\cite{zhao2007,chen2016,yin2017chinese,yin2017chinesezp}, metrics employed to evaluate our model are: recall, precision, and F-score (F). We report the performance for each source in addition to the overall result.

\subsubsection{Baselines and Experiment Settings}
Five recent zero pronoun resolution systems are employed as our baselines, namely, \newcite{zhao2007}, \newcite{chen2015}, \newcite{chen2016}, \newcite{yin2017chinese} and \newcite{yin2017chinesezp}. The first of them is machine learning-based, the second is the unsupervised and the other ones are all deep learning models. Since we concentrate on the anaphoric zero pronoun resolution process, we run experiments by employing the experiment setting with ground truth parse results and ground truth anaphoric zero pronoun, all of which are from the original dataset. Moreover, to illustrate the effectiveness of our reinforcement learning model, we run a set of ablation experiments by using different pretraining iterations and report the performance of our model with different iterations. Besides, to explore the randomness of the reinforcement learning technique, we report the performance variation of our model with different random seeds.
\subsubsection{Implementation Details}
We randomly initialize the parameters and minimize the objective function using Adagrad~\cite{duchi2011adaptive}. The embedding dimension is $100$, and hidden layers are $256$ and $512$ dimensions, respectively. Moreover, the dropout~\cite{hinton2012improving} regularization is added to the output of each layer. Table \ref{hyper} shows the hyperparameters we utilized for both the pre-training and reinforcement learning process.
       \begin{table}[h]
        \begin{center}
        \begin{tabular}{lcc}
        \Xhline{1pt}
        & Pre & RL \\
        \hline
        hidden dimentions & 256 \& 512 & 256 \& 512 \\
        training epochs & 70 & 50 \\
        batch & 256 & 256 \\
        dropout rate & 0.5 & 0.7 \\
        learning rate & 0.003 & 0.00009 \\
        \Xhline{1pt}
        \end{tabular}
        \end{center}
        \caption{Hyperparameters for the pre-training (Pre) and reinforcement learning (RL).}
        \label{hyper}
        \end{table}
        Hyperparameters here are selected based on preliminary experiments and there remains considerable space for improvement, for instance, applying the annealing.
            \begin{table*}
            \begin{center}
             \begin{tabular}{lcccccccc}
            \Xhline{1pt}

             & NW {\small(84)} & MZ {\small(162)} & WB {\small(284)} & BN {\small(390)} & BC {\small(510)} & TC {\small(283)} & & {\bf Overall} \\

            \Xhline{1pt}
            Zhao and Ng (2007) & 40.5 & 28.4 & 40.1 & 43.1  &  44.7 & 42.8 & & 41.5 \\
            Chen and Ng (2015) & 46.4 & 39.0 & 51.8 & 53.8 & 49.4 & 52.7 & & 50.2 \\
            Chen and Ng (2016) & 48.8 & 41.5 & 56.3 & 55.4 & 50.8 & 53.1 & & 52.2 \\
            Yin et al. (2017b) & 50.0 & 45.0 & 55.9 & 53.3  &  55.3 & 54.4 & & 53.6\\
            Yin et al. (2017a) & 48.8 & 46.3 & 59.8 & {\bf 58.4}  &  53.2 & {\bf 54.8} & & 54.9\\
            \hline
            {\bf Our model} & {\bf 63.1} & {\bf 50.2} & {\bf 63.1} & 56.7 & {\bf 57.5} & 54.0 & & {\bf 57.2} \\
            \Xhline{1pt}
            \end{tabular}
             \caption{\label{result_all}  Experiment results on the test data. The first six columns show the results on different source of documents and the last column is the overall results.}
            \end{center}
            \end{table*}

\subsection{Experiment Results}
In Table~\ref{result_all}, we compare the results of our model with baselines in the test dataset. Our reinforcement learning model surpasses all previous baselines. More specifically, for the ``Overall'' results, our model obtains a considerable improvement by $2.3\%$ in F-score over the best baseline~\cite{yin2017chinese}. Moreover, we run experiments in different sources of documents and report the results for each source. The number following a source's name indicates the amount of anaphoric zero pronoun in that source. Our model beats the best baseline in four of six sources, demonstrating the efficiency of our reinforcement learning model. The improvement gained over the best baseline in source ``BC''  is $4.3\%$ in F-score, which is encouraging since it contains the most anaphoric zero pronoun. In all words, all these suggest that our model surpasses existed baselines, which demonstrates the efficiency of the proposed technique.

Ideally, our model learns useful information gathered from candidates that have been predicted to be the antecedents in previous states, which brings a global-view instead of simply making partial decisions. By applying the reinforcement learning, our model learns to directly optimize the overall performance in expectation, guiding benefit in making decisions in a sequential manner. Consequently, they bring benefit to predict accurate antecedents, leading to the better performance.

Moreover, on purpose of better illustrating the effectiveness of the proposed reinforcement learning model, we run a set of experiments with different settings. In particular, we compare the model with and without the proposed reinforcement learning process using different pre-training iterations. For each time, we report the performance of our model on both the test and development set. For all these experiments, we retain the rest of the model unchanged.

        \begin{figure}[th]
            \centering
            \includegraphics[width=0.45\textwidth]{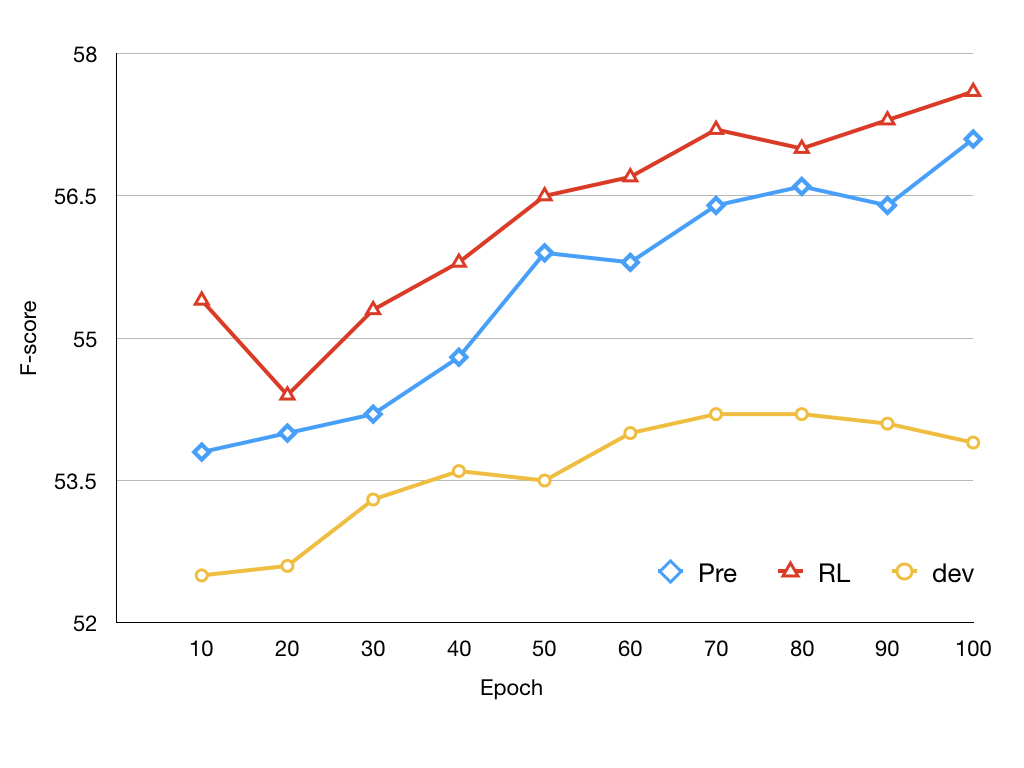}
            \caption{Experiment results of different models, where ``RL'' represents the reinforcement learning algorithm and ``Pre'' presents the model without reinforcement learning. ``dev'' shows the performance of our reinforcement learning model on the development dataset.}
            \label{tun}
        \end{figure}

Figure~\ref{tun} shows the performance of our model with and without reinforcement learning. We can see from the table that our model with reinforcement learning achieves better performance than the model without this all across the board. With the help of reinforcement learning, our model learns to choose effective actions in sequential decisions. It empowers the model to directly optimize the overall evaluation metrics, which brings a more effective and natural way of dealing with the task. Moreover, by seeing that the performance on development dataset stops increasing with iterations bigger than $70$, we therefore set the pre-training iterations to $70$.

Following~\newcite{reimers2017reporting}, to illustrate the impact of randomness in our reinforcement learning model, we run our model with different random seed values. Table~\ref{seed} shows the performance of our model with different random seeds on the test dataset. We report the minimum, the maximum, the median F-scores results and the standard deviation $\sigma$ of F-scores.
       \begin{table}[h]
        \begin{center}
        \begin{tabular}{cccc}
        \Xhline{1pt}
        Min F & Median F & Max F & $\sigma$ \\
        \hline
        56.5 & 57.1 &  57.5 & 0.00253\\
        \Xhline{1pt}
        \end{tabular}
        \end{center}
        \caption{Performance of our model with different random seeds.}
        \label{seed}
        \end{table}
We run the model with $38$ different random seeds. The maximum F-score is $57.5\%$ and the minimum one is $56.5\%$. Based on this observation, we can draw the conclusion that our proposed reinforcement learning model generally beats the baselines and achieves the state-of-the-art performance.

\subsection{Case Study}

        \begin{figure}[th]
            \centering
            \includegraphics[width=0.45\textwidth]{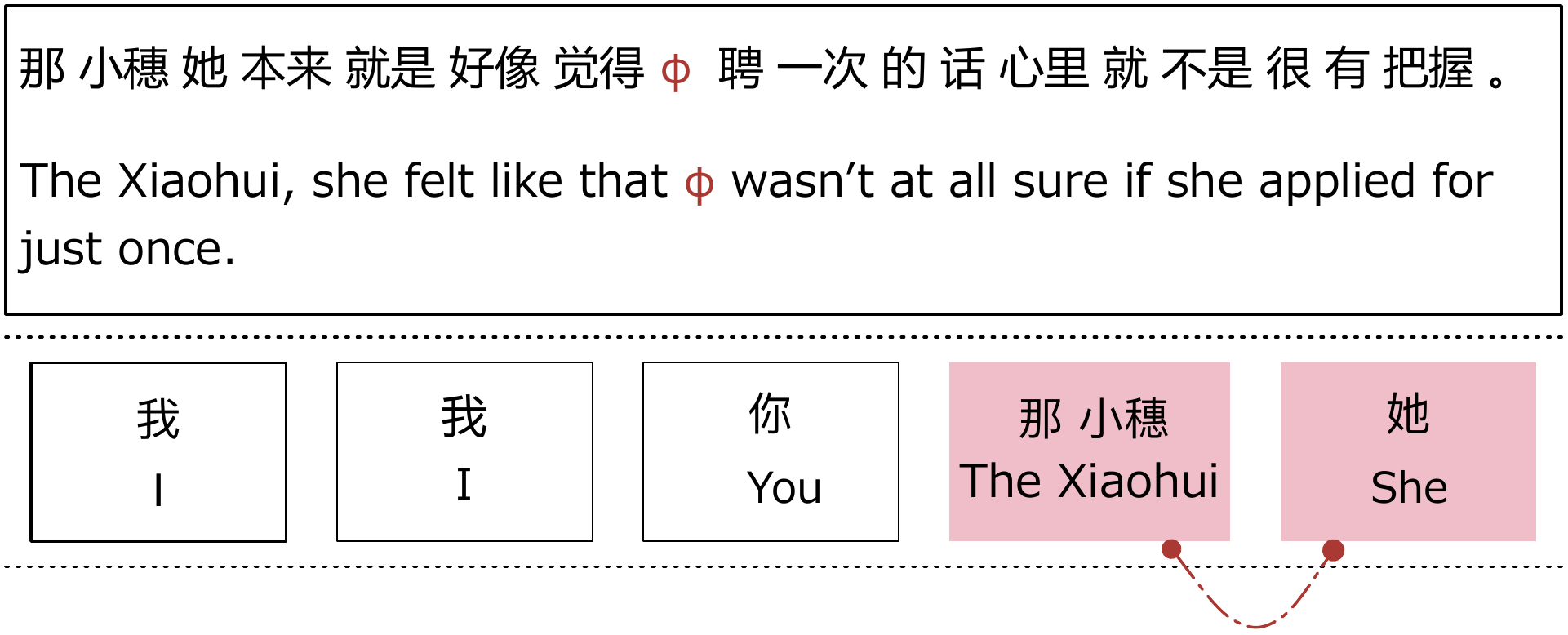}
            \caption{Example of case study. Noun phrases with pink background color are the ones selected to be the antecedents by our model.}
            \label{case}
        \end{figure}
Lastly, we show a case to illustrate the effectiveness of our proposed model, as is shown in Figure~\ref{case}. In this case, we can see that our model correctly predict mentions ``那 小穗/The Xiaohui'' and ``她/She'' as the antecedents of the zero pronoun ``$\phi$''. This case demonstrates the efficiency of our model. Instead of making only local decisions, our model learns to predict potential antecedents incrementally, selecting global-optimal antecedents in a sequential manner. In the end, our model successfully predicts ``她/She'' as the result.

\section{Related Work}
\label{sec:related_work}
\subsection{Zero Pronoun Resolution}
A wide variety of techniques for machine learning models for Chinese zero pronoun resolution have been proposed. \newcite{zhao2007} utilized the decision tree to learn the anaphoric zero pronoun resolver by using syntactical and positional features. It is the first time that machine learning techniques are applied for this task. To better explore syntactics, \newcite{kong2010} employed the tree kernel technique in their model. \newcite{chen2013} extended \newcite{zhao2007}'s model further by integrating innovative features and coreference chains between zero pronoun as bridges to find antecedents. In contrast, unsupervised techniques have been proposed and shown their efficiency. \newcite{chen2014} proposed an unsupervised model, where a model trained on manually resolved pronoun was employed for the resolution of zero pronoun. \newcite{chen2015} proposed an unsupervised anaphoric zero pronoun resolver, using the salience model to deal with the issue. Besides, there has been extensive work on zero anaphora for other languages. Efforts for zero pronoun resolution fall into two major categories, namely, (1) heuristic techniques~\cite{han2006korean}; and (2) learning-based models~\cite{iida2011cross,isozaki2003japanese,iida2006,iida2007zero,sasano2011,iida2011cross,yin2015joint,iida2015intra,iida2016intra}.

In recent years, deep learning techniques have been extensively studied for zero pronoun resolution. \newcite{chen2016} introduced a deep neural network resolver for this task. In their work, zero pronoun and candidates are encoded by a feed-forward neural network. \newcite{liu2016generating} explored to produce pseudo dataset for anaphoric zero pronoun resolution. They trained their deep learning model by adopting a two-step learning method that overcomes the discrepancy between the generated pseudo dataset and the real one. To better utilize vector-space semantics, \newcite{yin2017chinesezp} employed recurrent neural network to encode zero pronoun and antecedents. In particular, a two-layer antecedent encoder was employed to generate the hierarchical representation of antecedents. \newcite{yin2017chinese} developed an innovative deep memory network resolver, where zero pronouns are encoded by its potential antecedent mentions and associated text.

The major difference between our model and existed techniques lies in the applying of deep reinforcement learning. In this work, we formulate the anaphoric zero pronoun resolution as a sequential decision process in a reinforcement learning setting. With the help of reinforcement learning, our resolver learns to classify mentions in a sequential manner, making global-optimal decisions. Consequently, our model learns to take advantage of earlier predicted antecedents when making later coreference decisions.

\subsection{Deep Reinforcement Learning}
Recent advances in deep reinforcement learning have shown promise results in a variety of natural language processing tasks~\cite{branavan2012learning,narasimhan2015language,li2016deep}. In recent time, \newcite{clark2016deep} proposed a deep reinforcement learning model for coreference resolution, where an agent is utilized for linking mentions to their potential antecedents. They utilized the policy gradient algorithm to train the model and achieves better results compared with the counterpart neural network model. \newcite{narasimhan2016improving} introduced a deep Q-learning based slot-filling technique, where the agent's action is to retrieve or reconcile content from a new document. \newcite{xiong2017deeppath} proposed an innovative reinforcement learning framework for learning multi-hop relational paths. Deep reinforcement learning is a natural choice for tasks that require making incremental decisions. By combining non-linear function approximations with reinforcement learning, the deep reinforcement learning paradigm can integrate vector-space semantic into a robust joint learning and reasoning process. Moreover, by optimizing the policy-based on the reward signal, deep reinforcement learning model relies less on heuristic loss functions that require careful tuning.

\section{Conclusion}
\label{sec:conclude} 

We introduce a deep reinforcement learning framework for Chinese zero pronoun resolution. Our model learns the policy on selecting antecedents in a sequential manner, leveraging effective information provided by the earlier predicted antecedents. This strategy contributes to the predicting for later antecedents, bringing a natural view for the task. Experiments on the benchmark dataset show that our reinforcement learning model achieves an F-score of $67.2\%$ on the test dataset, surpassing all the existed models by a considerable margin.

In the future, we plan to explore neural network models for efficaciously resolving anaphoric zero pronoun documents and research on some specific components which might influence the performance of the model, such as the embedding. Meanwhile, we plan to research on the possibility of applying adversarial learning~\cite{goodfellow2014generative} to generate better rewards than the human-defined reward functions. Besides, to deal with the problematic scenario when ground truth parse tree and anaphoric zero pronoun are unavailable, we are interested in exploring the neural network model that integrates the anaphoric zero pronoun determination and anaphoric zero pronoun resolution jointly in a hierarchical architecture without using parser or anaphoric zero pronoun detector.

Our code is available at \url{https://github.com/qyyin/Reinforce4ZP.git}.

\section*{Acknowledgments}
Thank the anonymous reviewers for their valuable comments. This work was supported by the Major State Basic Research Development 973 Program of China (No.2014CB340503), National Natural Science Foundation of China (No.61472105 and No.61502120). According to the meaning by Harbin Institute of Technology, the contact author of this paper is Ting Liu.
\bibliography{acl2018}
\bibliographystyle{acl_natbib}

\end{CJK*}
\end{document}